\documentclass{article}

\PassOptionsToPackage{numbers, compress}{natbib}
\usepackage[preprint,nonatbib]{style}

\usepackage[utf8]{inputenc} 
\usepackage[T1]{fontenc}    
\usepackage{hyperref}       
\usepackage{url}            
\usepackage{booktabs}       
\usepackage{amsfonts}       
\usepackage{nicefrac}       
\usepackage{microtype}      

\usepackage{arydshln}
\usepackage{amssymb}
\usepackage{amsmath}
\usepackage{bm}
\usepackage{caption}
\usepackage{subcaption}

\usepackage[table]{xcolor}
\usepackage[export]{adjustbox}
\usepackage{wrapfig,lipsum,booktabs}
\usepackage{makecell}
\usepackage{mathrsfs}
\usepackage{multirow}
\usepackage{tcolorbox}

\newcommand{\cmark}{\ding{51}}
\newcommand{\xmark}{\ding{55}}
\usepackage{pifont}
\usepackage{enumitem}
\usepackage{fontawesome}
\usepackage{xspace}

\definecolor{mygray}{gray}{.9}
\definecolor{mygray1}{gray}{.97}
\definecolor{mygreen}{RGB}{93,173,85}

\makeatletter
\newcommand{\thickhline}{
    \noalign {\ifnum 0=`}\fi \hrule height 1pt
    \futurelet \reserved@a \@xhline
}

\newcommand{\model}{\textsc{PixelThink}\xspace}
\newcommand{\dataset}{ReasonSeg-\textsc{Diff}\xspace}

\definecolor{cpt_purple}{RGB}{166,82,166}
\definecolor{cpt_green}{RGB}{0,176,80}
\definecolor{cpt_yellow}{RGB}{229,157,35}
\definecolor{link}{RGB}{166,82,166}

\hypersetup{
  colorlinks=true,
  linkcolor=link,
  citecolor=link,
  urlcolor=link,
}

\usepackage{titletoc}
\usepackage[toc,page]{appendix}

\title{\model: \\Towards Efficient Chain-of-Pixel Reasoning}

\author{
Song Wang$^{1,2}$~ \ \ \  Gongfan Fang$^2$~ \ \ \ Lingdong Kong$^2$~ \ \ \  Xiangtai Li$^3$~  \ \ \ Jianyun Xu$^4\thanks{Project leader.}$ 
\\
\textbf{Sheng Yang}$^4$~ \ \ \ \textbf{Qiang Li}$^4$~ \ \ \ \textbf{Jianke Zhu}$^{1}\footnotemark[2]$~ \ \ \ \textbf{Xinchao Wang}$^{2}\thanks{Corresponding authors (\tt jkzhu@zju.edu.cn, xinchao@nus.edu.sg).}$
\\[1ex]
$^1$Zhejiang University \ \ \
$^2$National University of Singapore \ \ \ 
\\
$^3$Nanyang Technological University \ \ \
$^4$AD Lab, CaiNiao Inc., Alibaba Group \ \ \
\\[1ex]
\faGithubAlt~\textbf{Project Page:} \href{https://pixelthink.github.io}{\texttt{PixelThink.github.io}}
}

\begin{document}

\maketitle
\begin{figure}[h]
    \vspace{-0.7cm}
    \centering
    \includegraphics[width=\linewidth]{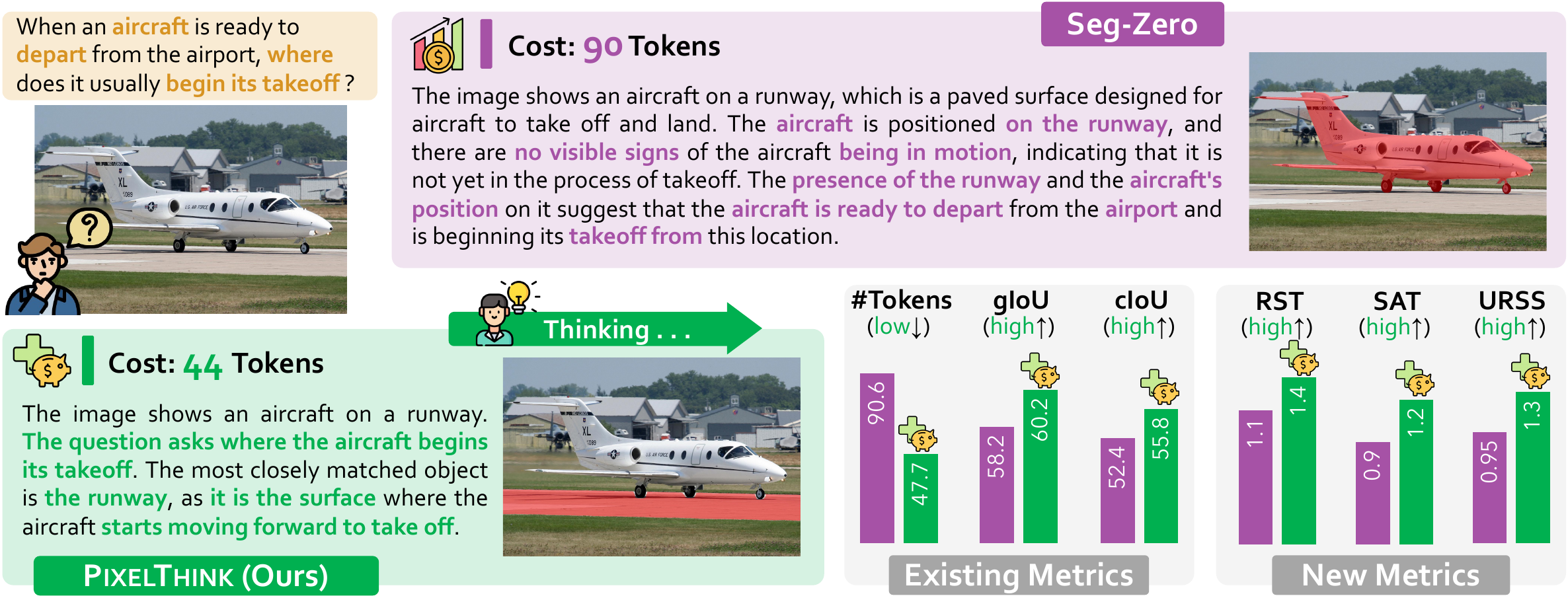}
    \vspace{-0.5cm}
    \caption{\textbf{Motivation of Efficient Chain-of-Pixel Reasoning (\model).} We propose a novel scheme for reasoning segmentation that effectively regulates reasoning length based on task \textit{difficulty} and \textit{uncertainty}. Our method improves segmentation quality while significantly reducing token usage. A suite of metrics is introduced for holistic evaluations of reasoning quality, segmentation accuracy, and computational efficiency.
    }
    \label{fig:teaser}
\end{figure}
\begin{abstract}
Existing reasoning segmentation approaches typically fine-tune multimodal large language models (MLLMs) using image-text pairs and corresponding mask labels. However, they exhibit limited generalization to out-of-distribution scenarios without an explicit reasoning process. Although recent efforts leverage reinforcement learning through group-relative policy optimization (GRPO) to enhance reasoning ability, they often suffer from overthinking -- producing uniformly verbose reasoning chains irrespective of task complexity. This results in elevated computational costs and limited control over reasoning quality. To address this problem, we propose \model, a simple yet effective scheme that integrates externally estimated task difficulty and internally measured model uncertainty to regulate reasoning generation within a reinforcement learning paradigm.  The model learns to compress reasoning length in accordance with scene complexity and predictive confidence. To support comprehensive evaluation, we introduce \dataset, an extended benchmark with annotated reasoning references and difficulty scores, along with a suite of metrics designed to assess segmentation accuracy, reasoning quality, and efficiency jointly. Experimental results demonstrate that the proposed approach improves both reasoning efficiency and overall segmentation performance. Our work contributes novel perspectives towards efficient and interpretable multimodal understanding. The code and model will be publicly available.
\end{abstract}

\section{Introduction}
\label{sec:intro}

Reasoning segmentation~\cite{yu2016refcoco, lai2024lisa, zhu2025popen} is an emerging vision-language task that requires predicting pixel-level masks in response to complex natural language queries. In contrast to traditional semantic or instance segmentation~\cite{chen2017deeplab, cheng2022masked, he2017mask}, which depends on predefined class labels, reasoning segmentation involves grounding fine-grained referring expressions that encode attributes, spatial relations, or contextual information. This capability is critical for embodied tasks such as interactive robotics~\cite{yin2023lamm, yu2025inst3d} and autonomous driving~\cite{tian2024drivevlm, xie2025vlms}. Advances in multimodal large language models (MLLMs)~\cite{liu2023llava,liu2024llava15, wang2024qwen2vl,bai2025qwen25} have facilitated the development of reasoning segmentation via supervised fine-tuning (SFT).

Representative approaches~\cite{lai2024lisa, ren2024pixellm, zhang2024omg-llava, bai2025onetokensegall}, such as the pioneering LISA~\cite{lai2024lisa}, integrate pre-trained MLLMs with segmentation modules through additional vision-language supervision to enable language-guided segmentation. Despite achieving strong performance on in-domain tasks, these methods often face limitations in generalizing to out-of-distribution (OOD) scenarios, especially when presented with complex or ambiguous queries~\cite{liu2025seg,shen2025vlm}. Moreover, the absence of explicit reasoning chains reduces interpretability and hinders effective error analysis.

To overcome the limitations of SFT-based methods, recent progress in LLM research~\cite{chen2025towards, li2025system} has motivated the adoption of reinforcement learning (RL) strategies to enhance reasoning capabilities and generalization. In particular, group-relative policy optimization (GRPO) has shown strong performance in language domains such as mathematical and code reasoning without requiring additional supervision~\cite{shao2024deepseekmath, guo2025deepseekr1}. Building on this, several works have extended GRPO to visual perception tasks~\cite{r1v, shen2025vlm, liu2025seg, liu2025visual}, achieving improved out-of-distribution generalization and generating explicit, interpretable reasoning paths. Nevertheless, these methods often suffer from overthinking -- producing unnecessarily verbose reasoning chains in simple cases -- which leads to increased computational cost and reduced efficiency~\cite{feng2025efficient, sui2025stop}. As illustrated in Figure~\ref{fig:teaser}, Seg-Zero~\cite{liu2025seg} yields incorrect segmentation results, despite utilizing a redundant reasoning process involving \textit{twice the number of tokens}. Moreover, the lack of standardized evaluation metrics for reasoning quality hinders a thorough assessment of the benefits brought by explicit reasoning in segmentation.

In this paper, we propose \model under the GRPO framework for reasoning segmentation, which regulates reasoning length based on externally estimated task difficulty and internally measured model uncertainty. Each input is assigned a token budget based on its estimated difficulty and uncertainty, and GRPO is guided by soft length-aware rewards that gently penalize excessive reasoning, promoting conciseness when appropriate. To facilitate systematic evaluation, we introduce \dataset, an extended version of ReasonSeg~\cite{lai2024lisa}, enriched with task difficulty annotations and dual-mode reasoning references (\textit{short} and \textit{long}). Our evaluation protocol jointly assesses segmentation accuracy (\texttt{gIoU} and \texttt{cIoU}) and reasoning score (\texttt{RScore}) using LLM-based ratings against reference reasoning chains. Additionally, we propose three efficiency-aware metrics to quantify the trade-off between segmentation performance and reasoning token usage. Specifically, \texttt{RST} and \texttt{SAT} assess the efficiency of reasoning and segmentation, respectively, while \texttt{URSS} provides a unified measure that comprehensively captures overall effectiveness across both aspects.

We conduct extensive experiments to benchmark \model against state-of-the-art reasoning segmentation and efficiency methods~\cite{liu2025seg,chen2024sam4mllm,lai2024lisa,aggarwal2025l1}. The results demonstrate that \model effectively regulates reasoning length in accordance with task difficulty, while maintaining reasoning quality and further enhancing segmentation accuracy. Additionally, we present exploratory analyses on the role of reasoning in segmentation, including ablations under no-thinking conditions. These analyses confirm that necessary and concise reasoning improves segmentation performance, whereas excessive redundancy provides no additional benefit.

Our main contributions can be summarized as follows:
\begin{itemize}
    \item We propose a novel scheme \model that enables efficient reasoning segmentation by leveraging external task difficulty and internal model uncertainty to guide the reward process in reinforcement fine-tuning.
    \item We build \dataset, a new benchmark with annotated reasoning references and difficulty scores, and establish a comprehensive evaluation protocol that covers reasoning quality, segmentation accuracy, and efficiency.
    \item Extensive experiments validate the effectiveness of our approach in reducing reasoning length and enhancing segmentation performance. In-depth analyses under various reasoning strategies are also conducted to inform future research.
\end{itemize}
\section{Related Work}
\label{sec:related_work}

\noindent \textbf{Reasoning Segmentation.} 
Referring expression segmentation~\cite{kazemzadeh2014referitgame,yu2016refcoco, wu2024towards, yang2024remamber, chng2024mask, huang2025mutual, pixel_sail} extends traditional segmentation approaches~\cite{ronneberger2015u,cheng2022masked,kirillov2023sam} to open-vocabulary settings by localizing objects described in natural language. Recent works leverage multimodal large language models (MLLMs)~\cite{liu2023llava, liu2024llava15, wang2024qwen2vl, bai2025qwen25} to integrate visual and linguistic reasoning, enabling flexible segmentation from free-form queries. LISA~\cite{lai2024lisa} introduces step-wise alignment between textual reasoning and object grounding. A series of subsequent works~\cite{ren2024pixellm, bai2025onetokensegall, zhu2025popen, yang2023lisa++, sa2va} explore fine-tuning MLLMs with segmentation heads, leveraging token-level instructions for fine-grained prediction. Seg-Zero~\cite{liu2025seg} further improves generalizability through reinforcement fine-tuning~\cite{guo2025deepseekr1,shao2024deepseekmath}, generating reasoning chains and reference tokens that guide segmentation modules. Despite these advances, current approaches still struggle with reasoning efficiency and adaptability to varying task difficulty levels. This work builds upon prior research and introduces an efficiency-aware reasoning scheme.

\noindent \textbf{Large Reasoning Models.}
Large language models (LLMs) have demonstrated remarkable capabilities in multi-step reasoning through Chain-of-Thought (CoT) prompting~\cite{wei2022chain, zhang2022automatic, yu2023towards}. Beyond prompting, recent efforts focus on optimizing the reasoning process via process reward models~\cite{wang2023mathshepherd, wang2025visualprm, song2025prmbench}, reinforcement fine-tuning~\cite{openaio1, team2025kimi, shao2024deepseekmath}, and other test time scaling methods~\cite{muennighoff2025s1,liu2025can, zuo2025ttrl}. DeepSeek-R1~\cite{guo2025deepseekr1} employs group relative policy optimization (GRPO)\cite{shao2024deepseekmath} to elicit LLMs' latent reasoning capacity and achieves significant advances. Building on this paradigm, recent works have extended GRPO and LLM-based reasoning to the visual domain~\cite{openr1multimodal, r1v, liu2025visual, tan2025reason, shen2025vlm}, enabling multimodal reasoning via vision-language models~\cite{wang2024qwen2vl, bai2025qwen25}. However, current visual reasoning research lacks evaluation of both the reasoning process and perceptual outcomes. We bridge this gap by introducing a holistic evaluation protocol that jointly assesses reasoning quality and segmentation performance.

\noindent \textbf{Efficient Inference Methods.}
Recent surveys \cite{feng2025efficient,sui2025stop, liu2025efficient, qu2025survey} have highlighted the inefficiencies of current reasoning models, including excessive token usage and redundant reasoning steps. To mitigate these issues, a variety of strategies have been explored. TALE~\cite{han2024token} proposes allocating token budgets adaptively, while CoT-Valve~\cite{ma2025cot} employs model merging to train reasoning chains of different lengths via supervised fine-tuning. Reinforcement learning-based methods such as L1~\cite{aggarwal2025l1} and O1-Pruner~\cite{luo2025o1} impose direct constraints on reasoning length during training. Further improvements include draft-based generation~\cite{xu2025chain}, skip mechanisms~\cite{xia2025tokenskip}, and pruning strategies~\cite{hou2025thinkprune}. Self-training approaches~\cite{munkhbat2025self} also contribute to efficiency by leveraging iterative pseudo-supervision. In the context of efficient MLLMs, efforts have primarily focused on dynamic and adaptive input compression~\cite{shang2024llava, li2024tokenpacker, chen2024image, yang2024visionzip,xu2025learning}, while output-level reasoning optimization remains underexplored. Our work addresses this problem by introducing a reward-driven mechanism that regulates reasoning length based on task difficulty and model uncertainty.
\section{Methodology}
\label{sec:method}

\subsection{Overview}

\noindent \textbf{Problem Definition.}
Reasoning segmentation~\cite{lai2024lisa, liu2025seg} aims to generate accurate segmentation masks given an image $\mathcal{I}$ and a referring expression $\mathcal{E}$.
In contrast to conventional referring segmentation, which directly maps inputs to segmentation outputs, our formulation additionally requires the model to produce an explicit intermediate reasoning process $\mathcal{R}$ to improve interpretability and generalization.
The task thus involves generating both the reasoning chain $\mathcal{R}$ and the segmentation mask $\mathcal{M}$.

\begin{figure}[t]
\begin{center}
\includegraphics[width=\linewidth]{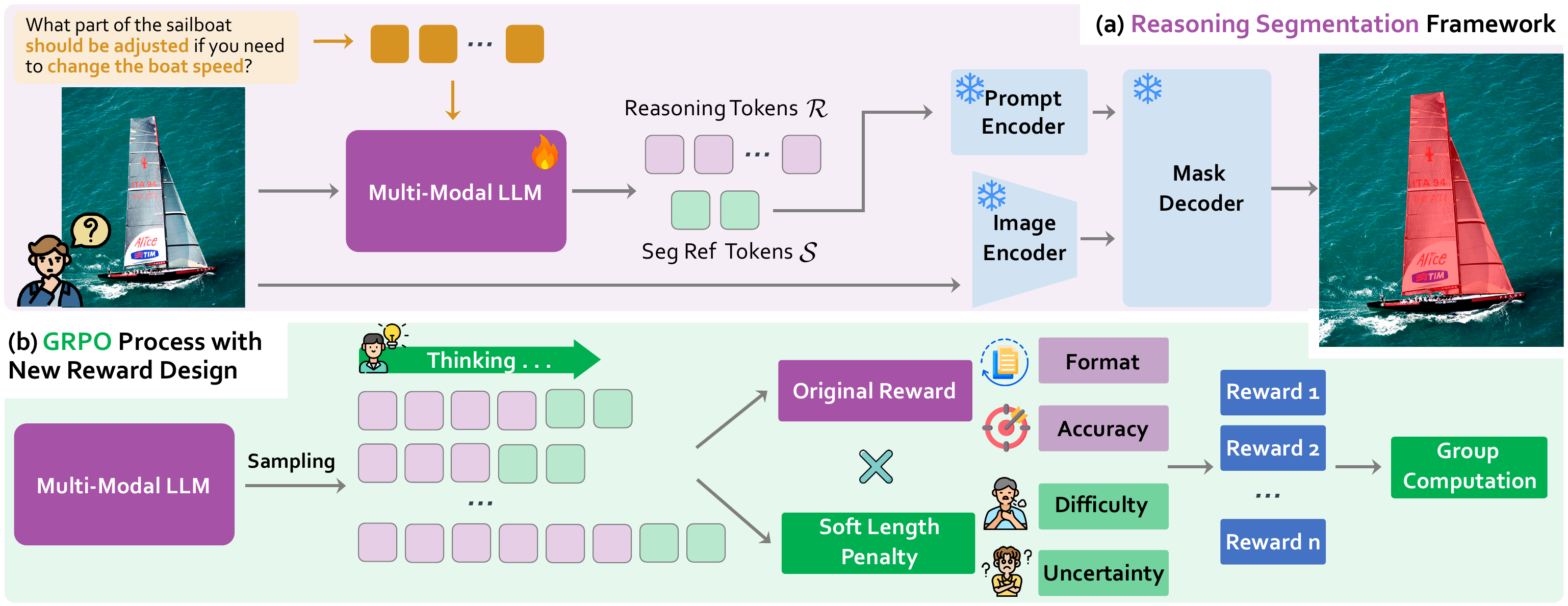} 
\end{center}
\vspace{-3mm}
\caption{\textbf{Overview of \model.} 
(a) Workflow of the reasoning segmentation framework. Given an input image and query, the model generates a reasoning chain and segmentation reference that guides the segmentation outcome. 
(b) The group-relative policy optimization (GRPO) procedure employed during reinforcement fine-tuning. 
Our new reward design incorporates both task difficulty and model uncertainty, enabling the model to learn efficient reasoning strategies.}
\label{fig:overallnetwork}
\end{figure}

\noindent \textbf{Baseline.}
We follow the standard setup~\cite{liu2025seg} as illustrated in Figure~\ref{fig:overallnetwork}(a), where the overall framework consists of a reasoning model and a segmentation model.
A multimodal large language model (MLLM), specifically Qwen2.5-VL~\cite{bai2025qwen25}, is adopted as the reasoning backbone (${\mathrm{Reason}}$). 
With an image $\mathcal{I}$ and a referring expression $\mathcal{E}$ as inputs, the model generates two distinct outputs:
$\mathcal{R}, \mathcal{S} = {\mathrm{Reason}}(\mathcal{I}, \mathcal{E})$,
where the reasoning chain $\mathcal{R}$ is a multi-step textual explanation that reflects the model's visual understanding and reasoning process. 
Segmentation reference tokens $\mathcal{S}$ are spatial priors including a bounding box and two points, which serve as inputs to the segmentation model.
We utilize SAM series models~\cite{kirillov2023sam, ravi2024sam2} as the segmentation module, which takes the segmentation reference token $\mathcal{S}$ predicted by the reasoning model as input and produces the final binary mask $\mathcal{M}$.
This modular design enables a clear decoupling of reasoning and fine-grained segmentation.

\noindent \textbf{Reinforcement Fine-Tuning.} 
We adopt reinforcement fine-tuning (RFT) to explicitly regulate output characteristics by optimizing non-differentiable objectives through task-specific reward signals. As shown in Figure~\ref{fig:overallnetwork}(b), our reward formulation integrates task difficulty and model uncertainty, promoting a balanced trade-off between reasoning quality and computational efficiency.

\subsection{Difficulty and Uncertainty Estimation}

\noindent \textbf{Task Difficulty.} To achieve efficient reasoning during training, we estimate an instance-level difficulty score $\mathcal{D} \in [1,10]$ for each sample. 
Following the same process in benchmark construction (Section~\ref{sec:dataset}), we prompt a large MLLM to assess difficulty across three aspects—\textit{scene complexity}, \textit{segmentation challenge}, and \textit{linguistic ambiguity}—and compute the final score as their average. 
These difficulty priors are then used to modulate token budget allocation in reinforcement fine-tuning.

\noindent \textbf{Model Uncertainty.}
We also quantify model-internal uncertainty based on token-level confidence in the generated reasoning sequence. For each token, we compute the gap between the highest and second-highest predicted probabilities~\cite{jiang2019minimum, wang2024chain}, using this margin to estimate certainty. The overall uncertainty score $\mathcal{U}$ is defined as:
\begin{equation}
\mathcal{U} = 1 - \frac{1}{T} \sum_{t=1}^{T} (p_{t}^{(1)} - p_{t}^{(2)}),
\end{equation}
where $p_{t}^{(1)}$ and $p_{t}^{(2)}$ are the top-2 probabilities at timestep $t$, and $T$ is the total number of tokens. 
A smaller margin indicates greater uncertainty, and the transformation ensures $\mathcal{U} \in [0,1]$, with higher values corresponding to lower confidence. 
This internal self-assessment complements external task difficulty, jointly informing the adjustment of reasoning length.

\subsection{Reward Design}
\label{sec:reward}
\noindent \textbf{Original Reward.}
We adopt the original reward design in Seg-Zero~\cite{liu2025seg}, which captures both reasoning validity and segmentation accuracy.
The reward function comprises the following components: 
\begin{equation}
\mathrm{R}_{\mathrm{original}} = \mathrm{R}_{\mathrm{format}}^{\mathrm{reason}} + \mathrm{R}_{\mathrm{format}}^{\mathrm{seg}} + 
\mathrm{R}_{\mathrm{accuracy}}^{\mathrm{seg}},
\end{equation}
which assesses reasoning format, segmentation format, and segmentation accuracy, respectively. $\mathrm{R}_{\mathrm{accuracy}}^{\mathrm{seg}}$ comprises the evaluation of mask IoU, point-level, and bounding box-level L1 distance.
The original reward $\mathrm{R}_{\mathrm{original}}$ serves as the basis for subsequent reasoning length modulation.

\noindent \textbf{Soft Length Penalty.}
To enable controllable reasoning length across tasks of varying complexity, we introduce a soft budget penalty that adaptively modulates the reward using both external task \textit{difficulty} and internal model \textit{uncertainty}. 
In contrast to prior approaches~\cite{aggarwal2025l1, han2024token}, this mechanism encourages concise reasoning in simple scenarios while permitting elaboration in complex or ambiguous cases, thereby mitigating overthinking and balancing reasoning adequacy with efficiency.
Given the difficulty score \( \mathcal{D}\) and the uncertainty score \( \mathcal{U}\), 
we define the expected reasoning token budget \( L_{\mathrm{budget}} \) as:
\begin{equation}
L_{\mathrm{budget}} = 
\begin{cases}
L_{\mathrm{base}} + \alpha \cdot \mathcal{U}, & \text{if } \mathcal{D} \geq \tau_1 \\
L_{\mathrm{low}}, & \text{if } \mathcal{D} < \tau_2 \\
\mathrm{None}, & \text{otherwise}
\end{cases}
\end{equation}
where thresholds \( \tau_1 \) and \( \tau_2 \) divide tasks into \textit{hard}, \textit{medium}, and \textit{easy} levels.
\( L_{\mathrm{base}} \), \( \alpha \), and \( L_{\mathrm{low}} \) are constants denoting the base budget for difficult tasks, the gain from uncertainty, and the minimal budget for easy tasks. We deliberately leave moderately difficult tasks unconstrained to allow learning flexibility in ambiguous regimes.
The soft penalty is computed as:
\begin{equation}
s(L_{\mathrm{used}}, L_{\mathrm{budget}}) = 
\begin{cases}
1 - \beta \cdot (L_{\mathrm{used}} - L_{\mathrm{budget}}), & \text{if } L_{\mathrm{used}} > L_{\mathrm{budget}} \\
1, & \text{otherwise}
\end{cases}
\end{equation}
where \( L_{\mathrm{used}} \) is the actual token count, and \( \beta \) is a small penalty factor that ensures smooth reward decay without harsh clipping. This design maintains model stability during training and avoids discouraging minor budget exceedance that may improve output quality.
The final reward is computed as:
\begin{equation}
\mathrm{R}_{\mathrm{final}} = \mathrm{R}_{\mathrm{original}} \cdot s(L_{\mathrm{used}}, L_{\mathrm{budget}}).
\end{equation}
This formulation achieves fine-grained adjustment over reasoning length, ensuring that token usage aligns with \textit{task complexity} and \textit{model confidence}. Moreover, it avoids the pitfalls of hard constraints, which could prematurely truncate informative reasoning during early training. 
Our empirical findings confirm that the chosen upper bounds are \textit{sufficiently permissive} to allow learning while providing enough structure to suppress redundant output.

\subsection{Reinforcement Fine-tuning Process}

\noindent\textbf{Training with GRPO.}
Following recent advances~\cite{shao2024deepseekmath, guo2025deepseekr1, liu2025seg}, we adopt group-relative policy optimization (GRPO) for reinforcement fine-tuning. 
The model is optimized to maximize the task-specific reward $\mathrm{R}_{\mathrm{final}}$ introduced in Section~\ref{sec:reward}, which integrates reasoning quality, segmentation accuracy, and token efficiency into a unified objective.
GRPO enhances training stability by comparing rewards within mini-batches at a group level, facilitating more consistent gradient updates and improving convergence with reduced variance.

\noindent\textbf{Inference.}
At inference time, the model operates under the same architecture and prompting schema as the baseline model. 
It generates both the reasoning chain $\mathcal{R}$ and segmentation mask $\mathcal{M}$ with image-text pairs as inputs. \textit{No additional labels} or \textit{reward feedback} are required during testing. 
This simple yet effective scheme facilitates seamless deployment and supports token-efficient, interpretable segmentation across diverse inputs.

\section{Benchmark Construction \& Evaluation Protocol}
\label{sec:benchmark}
While existing reasoning segmentation benchmarks primarily focus on evaluating the final segmentation mask, they often overlook the quality of the reasoning process and the efficiency of token usage. 
To address this gap and facilitate comprehensive evaluation, we construct \dataset, an extension of the ReasonSeg dataset~\cite{lai2024lisa} that includes task difficulty annotations and reference reasoning chains. 
This benchmark enables fine-grained assessment under varying levels of complexity.

\subsection{The Construction of \dataset}
\label{sec:dataset}
\begin{figure}[t]
    \centering
    \includegraphics[width=\linewidth]{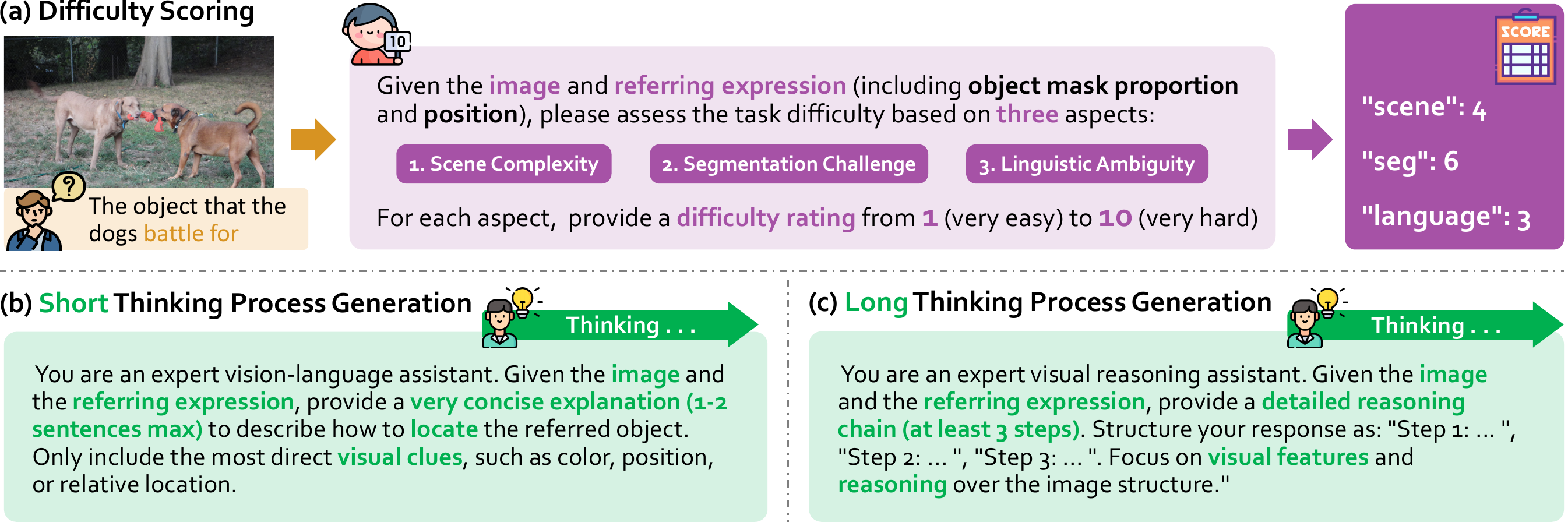} 
    \vspace{-0.5cm}
    \caption{\textbf{The construction of \dataset.} (a) Design on the \emph{Difficulty Scoring} scheme. (b) Generation of the \emph{Short Thinking} process. (c) Generation of \emph{Long Thinking} process. 
    }
\label{fig:overdataengine}
\end{figure}

As illustrated in Figure~\ref{fig:overdataengine}, the construction process consists of three parts, and we provide more details and examples in the \dataset Details part of the supplementary materials.

\noindent \textbf{Difficulty Scoring.}
To quantify the intrinsic reasoning challenge of each sample, we propose a structured scoring framework based on three interpretable factors: (1) \emph{Scene Complexity}, measuring the number and similarity of distractor objects; (2) \emph{Segmentation Challenge}, capturing the spatial size, position, occlusion, and whether the target is a whole or a part; and (3) \emph{Linguistic Ambiguity}, evaluating how explicit the referring expression is versus the need for visual inference.

As shown in Figure~\ref{fig:overdataengine}(a), we encode visual and textual priors into a unified prompt and query a large vision-language model (Qwen2.5-VL-72B~\cite{bai2025qwen25}) to independently score each aspect on a scale of $[1\text{–}10]$, accompanied by a natural language explanation. 
The final difficulty score is computed as the average of three dimensions, yielding a holistic and interpretable measure of instance-level complexity.
To ensure reliability, this score is further cross-validated against human annotations. We also categorize difficulty into three levels (\textit{easy}, \textit{medium}, and \textit{hard}) using thresholds $\tau_1$ and $\tau_2$.

\noindent \textbf{Short and Long Thinking Process Generation.}
To support evaluation under varying reasoning budgets, we construct two types of reference chains for each sample. 
The \emph{short chain} conveys essential visual cues and confident identification in $1$–$2$ sentences, while the \emph{long chain} follows a structured multi-step format (\textit{e.g.}, “Step $1$… Step $2$…”) to emulate more comprehensive visual reasoning.
Both chains are generated by prompting Qwen2.5-VL-72B with task-specific instructions and paired vision-text inputs as illustrated in Figure~\ref{fig:overdataengine}(b)(c). These only serve as evaluation references for assessing the informativeness and efficiency of model-generated reasoning.

\subsection{Evaluation Protocol}
\label{sec:eval}
With the constructed \dataset, we further introduce a comprehensive evaluation protocol for reasoning segmentation that jointly measures \emph{segmentation accuracy}, \emph{reasoning quality}, and \emph{computational efficiency}. The protocol consists of three complementary components detailed below.

\noindent \textbf{Segmentation Evaluation.}
Following established settings~\cite{yu2016refcoco,kazemzadeh2014referitgame}, we adopt two standard metrics: \texttt{gIoU}, the mean Intersection-over-Union (IoU) across all test samples, and \texttt{cIoU}, the cumulative IoU computed as the total intersection divided by the total union over the dataset.
To incorporate efficiency into segmentation assessment, we propose \textbf{Segmentation Accuracy per Token} (\texttt{SAT}), formulated as:
$
\texttt{SAT} = \frac{100 \times \texttt{gIoU}}{P \times \sqrt{T_{\mathrm{num}} + 1}},
$
where $P$ denotes the number of model parameters (in billions), and 
$T_{\mathrm{num}}$ is the average token length of the generated reasoning chains. 
The \textit{square root} in the denominator provides a soft penalty for longer outputs, discouraging unnecessarily verbose reasoning without overly punishing small increases in token length.
\texttt{SAT} favors models that achieve high segmentation accuracy with minimal reasoning overhead and compact model size.

\noindent\textbf{Reasoning Quality Evaluation.}
To assess the quality of generated reasoning chains, we employ a large language model (LLM)~\cite{yang2024qwenllm} to score predictions against the reference annotations introduced in Section~\ref{sec:dataset}. The evaluation covers three key dimensions: (1) \textit{Completeness} — whether all necessary steps and information are included; (2) \textit{Object Grounding} -- the degree of alignment with the referred visual object; and (3) \textit{Fluency and Clarity} -- coherence, grammaticality, and readability. 

Each aspect is rated on a $1$–$10$ scale, and their average constitutes the overall reasoning score, denoted as \texttt{RScore}.
To reflect reasoning efficiency, we further introduce \textbf{Reasoning Score per Token} (\texttt{RST}) as:
$
\texttt{RST} = \frac{10 \times \texttt{RScore}}{P \times \sqrt{T_{\mathrm{num}} + 1}}.
$
Following our difficulty-aware setting, \texttt{RScore} is computed using the \textit{short} chain for \emph{easy} and \emph{medium} samples, and the \textit{long} chain for \emph{hard} ones. This evaluation protocol favors models that generate concise yet semantically rich reasoning, especially with limited tokens.

\noindent\textbf{Unified Metric.}
To holistically evaluate reasoning segmentation across accuracy, reasoning quality, and computational efficiency, we propose the \textbf{Unified Reasoning Segmentation Score} (\texttt{URSS}):
$\texttt{URSS} = (1 - \gamma)  \times \texttt{RST} +  \gamma\times \texttt{SAT}$,
where $\gamma \in [0, 1]$ governs the relative emphasis on segmentation accuracy (\texttt{SAT}) and reasoning quality (\texttt{RST}). We set $\gamma = 0.7$ by default to reflect the \textit{greater importance} of segmentation performance in practical applications. 
\begin{table}[t]
    \centering
    \caption{Overall evaluation results on the proposed \dataset benchmark.}
    \vspace{0.1cm}
    \label{table:performance_comparison_reasoning_adapt}
    \resizebox{\textwidth}{!}{
    \begin{tabular}{l|ccc|ccc|c}
    \toprule[1.25pt]
    \rowcolor{gray!11} & \multicolumn{3}{c|}{\textbf{Reasoning Quality} } & \multicolumn{3}{c|}{\textbf{Segmentation Performance}} & \multicolumn{1}{c}{~\textbf{Overall}~}  
    \\
    \rowcolor{gray!11}\multicolumn{1}{c|}{\multirow{-2}{*}{\textbf{Method}}} & \multicolumn{1}{c}{{\#Token $\downarrow$}} & \multicolumn{1}{c}{{\texttt{RScore}$\uparrow$}} & \multicolumn{1}{c|}{{\texttt{RST}$\uparrow$}} &  \multicolumn{1}{c}{{\texttt{gIoU}{\small(\%)}$\uparrow$}}  & \multicolumn{1}{c}{{\texttt{cIoU}{\small(\%)}$\uparrow$}} & \multicolumn{1}{c|}{{\texttt{SAT}$\uparrow$}} & \multicolumn{1}{c}{{\texttt{URSS}$\uparrow$}}  
    \\\midrule\midrule
    \rowcolor{cpt_purple!15}\multicolumn{8}{c}{\em Source of ReasonSeg Data: Validation Set} 
    \\
    \textcolor{cpt_purple}{$\circ$} Seg-Zero~\cite{liu2025seg} & $90.79$  & ${7.67}$  & $1.14$  &  $61.63$ & $52.56$  & $0.92$ & $0.99$  
    \\
    \textcolor{cpt_purple}{$\circ$} Seg-Zero~\cite{liu2025seg} $+$ Prompt &  $57.62$ & $7.31$  & $1.36$  & $62.50$ & $59.22$  & $1.17$ & $1.23$   
    \\
    \textcolor{cpt_purple}{$\circ$} Seg-Zero~\cite{liu2025seg} $+$ L1-Exact~\cite{aggarwal2025l1}~ & $65.66$ & $5.73$  & $1.00$   & $40.97$  & $42.47$  & $0.72$ &$0.80$   
    \\
    \textcolor{cpt_purple}{$\circ$} Seg-Zero~\cite{liu2025seg} $+$ L1-Max~\cite{aggarwal2025l1} & $61.21$   & $4.37$  & $0.79$   &  $61.75$ & $57.39$  &  $1.12$ & $1.02$  
    \\
    \textcolor{cpt_purple}{$\bullet$} \textbf{\model} {\small({Ours})} & $\mathbf{46.98}$ & $6.92$ & $\mathbf{1.43}$ & $\mathbf{63.81}$ & $\mathbf{62.69}$ & $\mathbf{1.32}$ & $\mathbf{1.35}$   
    \\
    \midrule
    \rowcolor{cpt_green!15}
    \multicolumn{8}{c}{\em Source of ReasonSeg Data: Test Set} 
    \\
    \textcolor{cpt_green}{$\circ$} Seg-Zero~\cite{liu2025seg} & $90.58$  & ${7.67}$  & $1.14$  & $58.20$  & $52.37$  & $0.87$  & $0.95$
    \\
    \textcolor{cpt_green}{$\circ$} Seg-Zero~\cite{liu2025seg} $+$ Prompt & $57.84$  &  $7.33$ & $1.37$  & $58.15$ &  $53.45$ & $1.08$ & $1.17$  
    \\
    \textcolor{cpt_green}{$\circ$} Seg-Zero~\cite{liu2025seg} $+$ L1-Exact~\cite{aggarwal2025l1} & $65.24$  & $5.71$  & $1.00$  & $39.63$  & $34.42$  & $0.70$ & $0.79$   
    \\
    \textcolor{cpt_green}{$\circ$} Seg-Zero~\cite{liu2025seg} $+$ L1-Max~\cite{aggarwal2025l1} & $61.61$  & $4.31$  &  $0.78$ &  $58.20$ & $47.44$  & $1.05$ & $0.97$  
    \\
    \textcolor{cpt_green}{$\bullet$} \textbf{\model} {\small({Ours})} & $\mathbf{47.66}$  & $6.92$  & $\mathbf{1.42}$  & $\mathbf{60.17}$  & $\mathbf{55.77}$  &  $\mathbf{1.23}$ & $\mathbf{1.29}$  
    \\
    \bottomrule
\end{tabular}}
\vspace{-1mm}
\end{table}

\begin{table}[t]
    \caption{Difficulty-aware evaluation on the \dataset test set.}
    \vspace{0.1cm}
    \label{table:performance_comparison_reasoning_diff}
    \resizebox{\textwidth}{!}{
    \begin{tabular}{l|ccc|ccc|c}
    \toprule[1.25pt]
    \rowcolor{gray!11} & \multicolumn{3}{c|}{\textbf{Reasoning Quality} } & \multicolumn{3}{c|}{\textbf{Segmentation Performance}} & \multicolumn{1}{c}{~\textbf{Overall}~}  
    \\
    \rowcolor{gray!11}\multicolumn{1}{c|}{\multirow{-2}{*}{\textbf{Method}}} & \multicolumn{1}{c}{{\#Token $\downarrow$}} & \multicolumn{1}{c}{{\texttt{RScore}$\uparrow$}} & \multicolumn{1}{c|}{{\texttt{RST}$\uparrow$}} &  \multicolumn{1}{c}{{\texttt{gIoU}{\small(\%)}$\uparrow$}}  & \multicolumn{1}{c}{{\texttt{cIoU}{\small(\%)}$\uparrow$}} & \multicolumn{1}{c|}{{\texttt{SAT}$\uparrow$}} & \multicolumn{1}{c}{{\texttt{URSS}$\uparrow$}}  
    \\\midrule\midrule
    \rowcolor{cpt_purple!15}\multicolumn{8}{c}{\em Difficulty Level: Easy} 
    \\
    \textcolor{cpt_purple}{$\circ$} Seg-Zero~\cite{liu2025seg} & $84.97$  & ${8.07}$  &  $1.24$ &  $68.65$ & $65.85$  &  $1.06$ & $1.11$
    \\
    \textcolor{cpt_purple}{$\circ$} Seg-Zero~\cite{liu2025seg} $+$ Prompt &  $55.35$ & $7.80$  & $1.48$  & $67.93$  & $63.94$  &  $1.29$ & $1.35$  
    \\
    \textcolor{cpt_purple}{$\circ$} Seg-Zero~\cite{liu2025seg} $+$ L1-Exact~\cite{aggarwal2025l1}~ & $68.03$  & $6.42$ &   $1.10$ & $51.95$  &  $43.77$ &  $0.89$  & $0.96$
    \\
    \textcolor{cpt_purple}{$\circ$} Seg-Zero~\cite{liu2025seg} $+$ L1-Max~\cite{aggarwal2025l1} & $60.15$   &  $4.73$ & $0.86$  & $68.40$  &  $62.15$ &  $1.25$ & $1.13$
    \\
    \textcolor{cpt_purple}{$\bullet$} \textbf{\model} {\small({Ours})} &  $\mathbf{44.73}$ & $7.56$  & $\mathbf{1.60}$  & $\mathbf{70.25}$  & $\mathbf{67.49}$  & $\mathbf{1.48}$ & $\mathbf{1.52}$   
    \\
    \midrule
    \rowcolor{cpt_green!15}
    \multicolumn{8}{c}{\em Difficulty Level: Medium} 
    \\
    \textcolor{cpt_green}{$\circ$} Seg-Zero~\cite{liu2025seg} & $90.73$ &  ${7.68}$ & $1.15$  & $60.15$  & $55.53$  &   $0.90$ & $0.97$ 
    \\
    \textcolor{cpt_green}{$\circ$} Seg-Zero~\cite{liu2025seg} $+$ Prompt & $58.37$  & $7.30$  &  $1.35$ & $59.89$ & $56.15$  & $1.11$ &  $1.18$  
    \\
    \textcolor{cpt_green}{$\circ$} Seg-Zero~\cite{liu2025seg} $+$ L1-Exact~\cite{aggarwal2025l1} & $66.03$  & $5.60$  &  $0.98$ &  $39.97$ &  $35.01$ &  $0.70$  & $0.78$
    \\
    \textcolor{cpt_green}{$\circ$} Seg-Zero~\cite{liu2025seg} $+$ L1-Max~\cite{aggarwal2025l1} & $61.45$  & $4.18$  & $0.76$  & $59.46$  & $47.79$  &  $1.07$ & $0.98$  
    \\
    \textcolor{cpt_green}{$\bullet$}  \textbf{\model} {\small({Ours})} & $\mathbf{47.00}$  & $6.84$  & $\mathbf{1.41}$  &  $\mathbf{62.05}$ &  $\mathbf{58.16}$ & $\mathbf{1.28}$ & $\mathbf{1.32}$   
    \\
    \midrule
    \rowcolor{cpt_yellow!15}\multicolumn{8}{c}{\em Difficulty Level: Hard} 
    \\
    \textcolor{cpt_yellow}{$\circ$} Seg-Zero~\cite{liu2025seg} & $95.37$  &  ${7.26}$ & $1.06$   &  $44.37$ & $28.93$  & $0.65$ & $0.77$   
    \\
    \textcolor{cpt_yellow}{$\circ$} Seg-Zero~\cite{liu2025seg} $+$ Prompt & $58.97$  &  $6.97$ &  $\mathbf{1.29}$ & $45.38$  &  $31.35$  & $0.84$ & $0.97$  
    \\
    \textcolor{cpt_yellow}{$\circ$} Seg-Zero~\cite{liu2025seg} $+$ L1-Exact~\cite{aggarwal2025l1} & $60.94$  &  $5.29$ & $0.96$   &  $27.61$ & $18.94$  & $0.50$ & $0.64$   
    \\
    \textcolor{cpt_yellow}{$\circ$} Seg-Zero~\cite{liu2025seg} $+$ L1-Max~\cite{aggarwal2025l1} & $63.28$  &  $4.22$ & $0.75$   & $46.09$  & $30.36$  &  $0.82$ & $0.80$  
    \\
    \textcolor{cpt_yellow}{$\bullet$} \textbf{\model} {\small({Ours})} & $\mathbf{51.79}$  & $6.50$  & $1.28$  &  $\mathbf{46.80}$ & $\mathbf{35.05}$  & $\mathbf{0.92}$ & $\mathbf{1.03}$   
    \\
    \bottomrule
    \end{tabular}
    }
    \vspace{-1mm}
\end{table}

\section{Experiments}
\label{sec:exps}

\noindent \textbf{Datasets.}  
We train exclusively on $9,000$ samples from RefCOCOg~\cite{yu2016refcoco} \textit{without any reasoning data}, following the same split as Seg-Zero~\cite{liu2025seg} for fair comparison. 
Evaluation is primarily conducted on \dataset derived from ReasonSeg~\cite{lai2024lisa}, enabling \textit{zero-shot} assessment across varying task complexities. \dataset includes $199$ validation samples and $769$ test samples, stratified by difficulty into $51$/$102$/$45$ and $171$/$411$/$187$ for easy, medium, and hard levels, respectively.
We further report results on RefCOCO, RefCOCO+, and RefCOCOg to validate the common performance.

\noindent \textbf{Implementation Details.}  
We mainly initialize the reasoning model with Qwen2.5-VL-7B~\cite{bai2025qwen25} and adopt SAM2-Large~\cite{ravi2024sam2} as the segmentation backbone. Reinforcement fine-tuning is performed using the GRPO~\cite{shao2024deepseekmath}, with a KL divergence coefficient of $1\times10^{-3}$ and $8$ samples per update. The initial learning rate is set to $1\times10^{-6}$.
The soft length penalty parameters are set as follows: $L_{\mathrm{base}} = 256$, $\alpha = 25$, $L_{\mathrm{low}} = 96$, and the penalty factor $\beta = 2 \times 10^{-3}$.
The thresholds $\tau_{1}$ and $\tau_{2}$ are set to $5.0$ and $3.5$, respectively.
All experiments are conducted on $8$ NVIDIA A100 GPUs with DeepSpeed~\cite{rasley2020deepspeed}.

\subsection{Main Results}

\noindent \textbf{Quantitative Results on \dataset.}
We evaluate our method on our \dataset benchmark, which incorporates fine-grained difficulty annotations and reference reasoning chains to facilitate comprehensive assessment across reasoning quality, segmentation accuracy, and efficiency. 
As Seg-Zero~\cite{liu2025seg} is the only existing approach that jointly provides explicit reasoning chains and segmentation masks, we adopt it as the primary baseline for comparison.
To examine efficient reasoning generation, we additionally compare with three length-aware baselines: (1) \textbf{Prompt}, which imposes a token limit through prompt-level constraints; (2) \textbf{L1-Exact}, and (3) \textbf{L1-Max}, two reinforcement fine-tuning strategies from~\cite{aggarwal2025l1} that incorporate different reward formulations to regulate output length. 
For fair comparison with Seg-Zero, all methods are fine-tuned exclusively on the RefCOCOg~\cite{yu2016refcoco} training split without any additional annotations.

Table~\ref{table:performance_comparison_reasoning_adapt} presents the overall results on the validation and test sets of \dataset. Our method achieves substantial reductions in reasoning token usage while simultaneously enhancing segmentation accuracy. 
Unlike prompt-based and L1-style baselines that enforce rigid length constraints, our approach preserves reasoning quality, resulting in a \textit{more favorable balance} between performance and efficiency. Additionally, Table~\ref{table:performance_comparison_reasoning_diff} reports difficulty-level breakdowns on the test set, where our method consistently surpasses all baselines across easy, medium, and hard subsets.

\begin{table}[t]
\centering
\caption{Performance comparison on existing benchmarks. Symbol $^{\dagger}$ denotes scores reported in the paper while $^{*}$ denotes our reproduction with official code and model checkpoint. 
Our method achieves consistent improvements across the majority of benchmarks. 
The 7B model shows limited gains on RefCOCO due to the dataset’s inherent simplicity and performance saturation.
}
\label{tab:pre_bench}
\vspace{-0.1cm}
\noindent
\begin{minipage}[t]{0.49\textwidth}
\centering
\caption*{(a) Reasoning segmentation on ReasonSeg~\cite{lai2024lisa}.}
\adjustbox{max width=\linewidth}{
\begin{tabular}{l|cc|cc}
\toprule[1.25pt]
\rowcolor{gray!11} & \multicolumn{2}{c|}{\texttt{val}} & \multicolumn{2}{c}{\texttt{test}} 
\\
\rowcolor{gray!11}\multicolumn{1}{c|}{\multirow{-2}{*}{Method}}  & \texttt{gIoU} & \texttt{cIoU} &  \texttt{gIoU} & \texttt{cIoU} 
\\
\midrule\midrule
OVSeg~\cite{liang2023ovseg} & $28.5$ & $18.6$ & $26.1$ & $20.8$  
\\
ReLA~\cite{liu2023rela}       & $22.4$ & $19.9$ & $21.3$ & $22.0$  
\\
Grounded-SAM~\cite{ren2024groundedsam} & $26.0$ & $14.5$ & $21.3$ & $16.4$ 
\\
LISA-7B-LLaVA1.5~\cite{lai2024lisa} & $53.6$ & $52.3$ & $48.7$ & $48.8$ 
\\
LISA-13B-LLaVA1.5~\cite{lai2024lisa} & $57.7$ & $60.3$ & $53.8$ & $50.8$ 
\\
SAM4MLLM~\cite{chen2024sam4mllm}   & $46.7$ & $48.1$ & - & - 
\\
Qwen2.5VL-3B~\cite{bai2025qwen25} $+$ SAM2~\cite{ravi2024sam2} & $53.8$ & $44.1$ & $47.6$ & $37.4$ 
\\
{Seg-Zero-3B$^{\dagger}$~\cite{liu2025seg}} &  {$62.6$} &  {$58.5$} &  {$56.1$} &  {$48.6$} 
\\ 
{Seg-Zero-7B$^{\dagger}$~\cite{liu2025seg}} &  {{$62.6$}} &  {{$62.0$}} &  {{$57.5$}} &  {{$52.0$}} \\ 
\midrule
{Seg-Zero-3B$^{*}$~\cite{liu2025seg}} &  $59.1$ & $48.8$  &  $52.5$ &  $43.4$
\\ 
{Seg-Zero-7B$^{*}$~\cite{liu2025seg}} & $61.6$  &  $52.6$ & $58.2$  & $52.4$  \\ 
{\textbf{\model}-3B (ours)} &  {$62.3$} &  {$58.5$} &  {$58.8$} &  {$52.1$} \\
{\textbf{\model}-7B (ours)} &  {$\mathbf{63.8}$} &  {$\mathbf{62.7}$} &  {$\mathbf{60.2}$} &  {$\mathbf{55.8}$} 
\\
\bottomrule
\end{tabular}
}
\end{minipage}
\hfill
\begin{minipage}[t]{0.485\textwidth}
\centering
\caption*{(b) Referring expression segmentation (cIoU) on RefCOCO (+/g)~\cite{yu2016refcoco}.}
\adjustbox{max width=\linewidth}{
\begin{tabular}{l|ccc}
\toprule[1.25pt]
\rowcolor{gray!11} & \multicolumn{1}{c}{\textbf{RefCOCO}} & \multicolumn{1}{c}{\textbf{RefCOCO+}} & \multicolumn{1}{c}{\textbf{RefCOCOg}} 
\\
\rowcolor{gray!11}\multicolumn{1}{c|}{\multirow{-2}{*}{Method}} & \texttt{testA} & \texttt{testA} &  \texttt{test} 
\\
\midrule\midrule
LAVT~\cite{yang2022lavt}               & $75.8$  & $68.4$ &  $62.1$ 
\\
ReLA~\cite{liu2023rela}               & $76.5$  & $71.0$ &  $66.0$ 
\\
LISA-7B~\cite{lai2024lisa}            & $76.5$  & $67.4$ &  $68.5$ 
\\
PixelLM-7B~\cite{ren2024pixellm}         & $76.5$  & $71.7$ &  $70.5$ 
\\
MagNet~\cite{chng2024mask} & $78.3$  & $73.6$ &  $69.3$ 
\\
PerceptionGPT-7B~\cite{pi2024perceptiongpt}   & $78.6$  & $73.9$ &  $71.7$ 
\\
{Seg-Zero-3B$^{\dagger}$~\cite{liu2025seg}} &  {$79.3$}  &   {$73.7$}&  {$71.5$}
\\
{Seg-Zero-7B$^{\dagger}$~\cite{liu2025seg}} &  {$80.3$}  &  {$76.2$} &   {$72.6$}
\\ 
\midrule
{Seg-Zero-3B$^{*}$~\cite{liu2025seg}} &  $76.0$  & $70.6$  & $68.8$ 
\\ 
{Seg-Zero-7B$^{*}$~\cite{liu2025seg}} & ${79.4}$   & $73.7$  & $73.2$  
\\ 
{\textbf{\model}-3B (ours)} &  {$78.7$}  &   {$72.9$}&  {$72.2$}
\\ 
{\textbf{\model}-7B (ours)} &  {${79.3}$}  &  {$\mathbf{74.8}$} &   {$\mathbf{73.9}$}
\\
\bottomrule
\end{tabular}
}
\end{minipage}
\vspace{-2mm}
\end{table}
\begin{figure}[t]
    \vspace{-5mm}
    \centering
    \includegraphics[width=\linewidth]{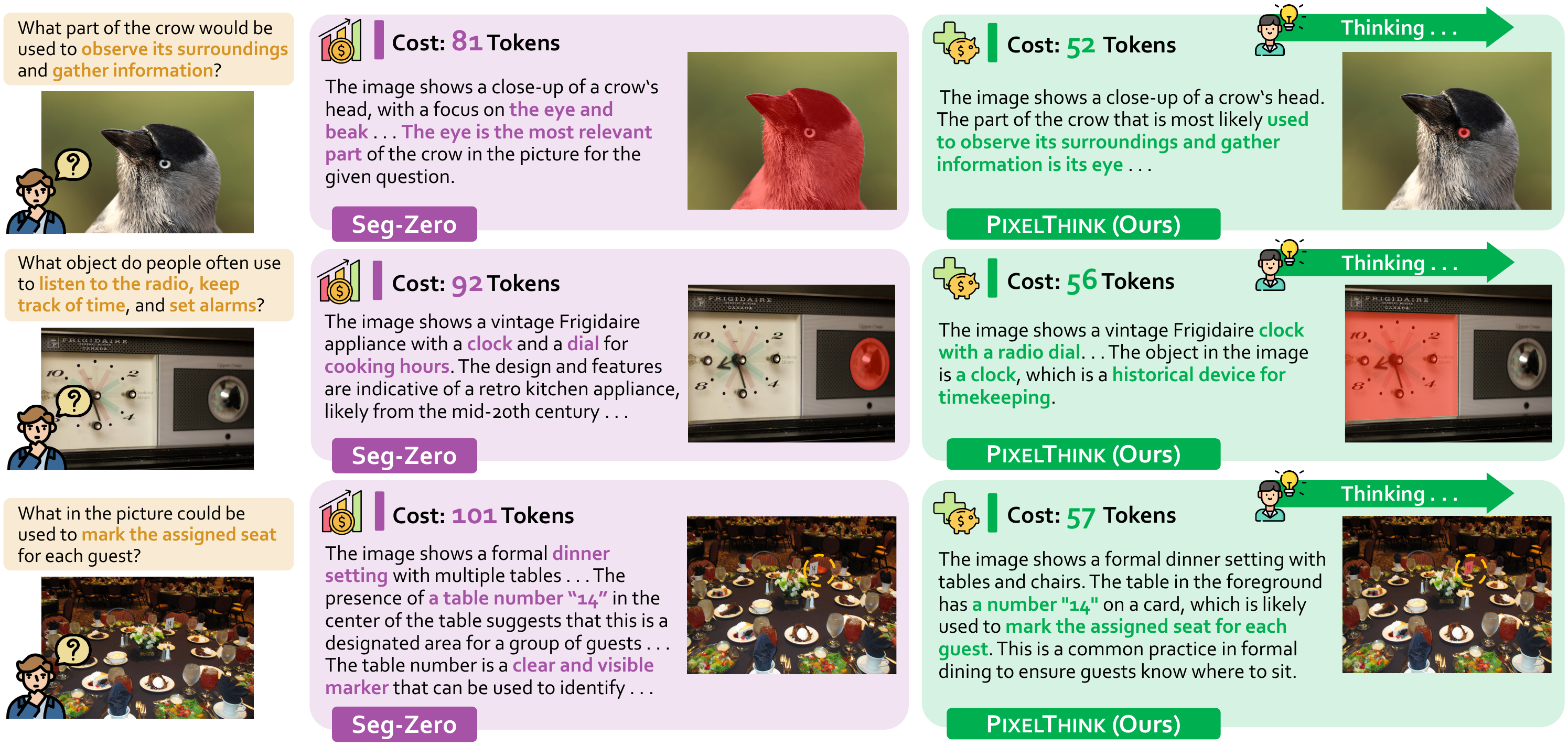}
    \caption{\textbf{Qualitative comparisons} between Seg-Zero~\cite{liu2025seg} and the proposed \model. Representative samples across different difficulty levels are selected to highlight differences in the reasoning process and segmentation performance.
    }
    \label{fig:vis}
    \vspace{-2mm}
\end{figure}

\noindent \textbf{Quantitative Results on Existing Benchmark.}
We further evaluate the generalization capability of our method on four widely-used benchmarks: ReasonSeg~\cite{lai2024lisa} and the standard referring expression segmentation datasets RefCOCO, RefCOCO+, and RefCOCOg~\cite{yu2016refcoco}.
On ReasonSeg, we compare against state-of-the-art approaches including OVSeg~\cite{liang2023ovseg}, ReLA~\cite{liu2023rela}, LISA~\cite{lai2024lisa}, Grounded-SAM~\cite{ren2024groundedsam}, and SAM4MLLM~\cite{chen2024sam4mllm}. As reported in Table~\ref{tab:pre_bench}(a), our method achieves the highest segmentation accuracy while using significantly fewer reasoning tokens. Notably, our method with Qwen2.5-VL-3B model performs on par with prior 7B counterparts, highlighting the efficiency of our approach.

We also report results on RefCOCO, RefCOCO+, and RefCOCOg using their standard test splits. As shown in Table~\ref{tab:pre_bench}(b), our method delivers competitive performance across all three datasets, confirming its robustness and strong generalization in both in-domain and out-of-domain settings.

\noindent \textbf{Qualitative Results.} 
Figure~\ref{fig:vis} presents visual comparisons between our method and Seg-Zero across various scenes. \model consistently predicts more precise segmentation masks while generating substantially shorter reasoning chains. 
In contrast, Seg-Zero frequently exhibits \textit{overthinking}, producing lengthy and redundant explanations that fail to improve segmentation quality.

\subsection{Diagnostic Experiments}
We conduct diagnostic experiments on the validation set of \dataset for further exploration. Additional results and implementation details are provided in the supplementary materials.

\begin{wraptable}{r}{0.48\linewidth}
    \centering
    \setlength{\abovecaptionskip}{0cm}
    \captionsetup{width=.48\textwidth}
    \caption{Ablation of our \model scheme.}
    \label{tab:abl_framework}
    \resizebox{0.48\textwidth}{!}{
    \begin{tabular}{cc|ccc}
    \toprule[1.25pt]
    \rowcolor{gray!11}Difficulty & Uncertainty & \#Token $\downarrow$  & \texttt{gIoU}(\%)$\uparrow$ & \texttt{cIoU}(\%)$\uparrow$ 
    \\\midrule\midrule
    &  &  $86.92$    & $59.65$  &  $50.92$ 
    \\
    \cmark &  &  $\mathbf{39.95}$   & $62.07$ & $58.35$  
    \\
    & \cmark & $42.41$     & $59.54$ &  $52.37$  
    \\
    \cmark &\cmark &   $46.98$   & $\mathbf{63.81}$ & $\mathbf{62.69}$  
    \\
    \bottomrule
    \end{tabular}}
    \vspace{-4mm}
\end{wraptable}
\noindent \textbf{Ablation on \model Scheme.} 
We ablate the two central components of \model -- \textit{task difficulty} and \textit{model uncertainty} -- to analyze their individual and joint contributions. 
For uncertainty-guided control only, we also divide the uncertainty scores into three levels and assign token budgets accordingly. 
As shown in Table~\ref{tab:abl_framework}, incorporating either difficulty or uncertainty alone reduces reasoning length and yields improvements in segmentation accuracy. 
Their combination achieves the optimal performance, confirming the complementary nature of the two signals for effective and efficient reasoning.

\begin{table}[t]
    \centering
    \begin{minipage}[t]{0.3\textwidth}
    \centering
    \caption{Ablation of the difficulty splits in \model.}
    \vspace{0.1cm}
    \label{tab:abl_diff}
    \resizebox{\textwidth}{!}{
    \begin{tabular}{c|ccc}
    \toprule[1.25pt]
    \rowcolor{gray!11}Difficulty & \#Tok$\downarrow$ & \texttt{gIoU}$\uparrow$ & \texttt{cIoU}$\uparrow$ 
    \\
    \midrule\midrule
    - & $86.92$ & $59.65$ & $50.92$ 
    \\
    2 & $61.78$ & $\mathbf{64.38}$ & $61.80$ 
    \\
    3 & $\mathbf{46.98}$ & $63.81$ & $\mathbf{62.69}$ 
    \\
    4 & $76.89$ & $62.28$ & $55.78$ 
    \\
    \bottomrule
    \end{tabular}
    }
    \end{minipage}
    \hfill
    \begin{minipage}[t]{0.3\textwidth}
    \centering
    \caption{Ablation on the token budget allocation details.}
    \vspace{0.1cm}
    \label{tab:abl_token}
    \resizebox{\textwidth}{!}{
    \begin{tabular}{c|ccc}
    \toprule[1.25pt]
    \rowcolor{gray!11}Budget & \#Tok$\downarrow$  & \texttt{gIoU}$\uparrow$ & \texttt{cIoU}$\uparrow$ 
    \\
    \midrule\midrule
    -  &  $86.92$ & $59.65$ & $50.92$   
    \\
    ($64$, $256$) & $\mathbf{24.71}$ & $62.12$ & $59.57$ 
    \\
    ($96$, $256$) & $46.98$ & $\mathbf{63.81}$ & $\mathbf{62.69}$ 
    \\
    ($96$, $384$) & $60.18$ & $60.18$ & $53.29$ 
    \\
    \bottomrule
    \end{tabular}
    }
    \end{minipage}
    \hfill
    \begin{minipage}[t]{0.36\textwidth}
    \centering
    \caption{Ablation on the different no-thinking mode analyses.}
    \vspace{0.1cm}
    \label{tab:abl_nothinking}
    \resizebox{\textwidth}{!}{
    \begin{tabular}{c|ccc}
    \toprule[1.25pt]
    \rowcolor{gray!11}Method & \#Tok$\downarrow$  & \texttt{gIoU}$\uparrow$ & \texttt{cIoU}$\uparrow$ 
    \\\midrule\midrule
    No-thinking-RL & $0.00$   & $60.19$   & $49.49$  
    \\
    No-thinking-Prompt  & $0.00$  &  $60.37$ &  $51.47$ 
    \\
    Seg-Zero  &  $90.79$  &  $61.63$   &  $52.56$ 
    \\
    \textbf{\model}(ours)  &  $46.98$  &  $\mathbf{63.81}$ &  $\mathbf{62.69}$
    \\
    \bottomrule
    \end{tabular}
    }
    \end{minipage}
    \vspace{-2mm}
\end{table}

\noindent \textbf{Ablation on Difficulty Splits.} 
We further investigate the impact of difficulty granularity by varying the number of difficulty levels used during training. 
In the $2$-level setting, medium and hard samples are merged and assigned a uniformly larger token budget. 
In the $4$-level setting, the medium category is further subdivided, with finer-grained length constraints applied. 
As illustrated in Table~\ref{tab:abl_diff}, the $3$-level split offers the best balance between segmentation accuracy and reasoning efficiency.

\noindent \textbf{Ablation on Token Budget Allocation.} 
We next ablate the token budget configuration used for the soft length penalty in reward. 
Specifically, we set the base upper bounds to $L_{\mathrm{base}} = 256$ and $L_{\mathrm{low}} = 96$, and assess alternative parameter configurations accordingly.
As shown in Table~\ref{tab:abl_token}, our approach consistently achieves substantial reductions in reasoning token usage while improving segmentation performance, demonstrating the robustness of the proposed budget design.

\noindent \textbf{No-thinking Mode Analyses.}
Inspired by recent investigations into no-thinking paradigms for reasoning~\cite{li2025cls, ma2025reasoning}, we extend this line of analysis to the pixel-level segmentation task. 
We implement two variants within the reinforcement learning framework: No-thinking-RL and No-thinking-Prompt. 
As demonstrated in Table~\ref{tab:abl_nothinking}, incorporating appropriate reasoning steps consistently improves segmentation accuracy, highlighting the benefit of efficient reasoning over naive or omitted inference.
\section{Conclusion}
\label{sec:conclusion}

In this paper, we propose \model, an efficiency-aware reasoning scheme for segmentation that explicitly regulates reasoning length based on task difficulty and model uncertainty. 
By introducing a soft length penalty and reward modulation, our method enables efficient chain-of-Pixel reasoning and improving segmentation accuracy. 
To achieve comprehensive evaluation, we construct a difficulty-aware benchmark \dataset, and design holistic metrics that jointly assess reasoning quality, segmentation precision, and efficiency. 
Extensive experiments demonstrate that \model produces concise yet informative reasoning chains and consistently outperforms baselines across varying difficulty levels. 
Further discussions are presented in Section~\ref{sec:supp_dis} of the Appendix.

\appendix
\setcounter{figure}{0}
\setcounter{table}{0}
\renewcommand{\thefigure}{\Alph{figure}}
\renewcommand{\thetable}{\Alph{table}}

\section*{Appendix}
\vspace{-0.2cm}
\startcontents[appendices]
\printcontents[appendices]{l}{1}{\setcounter{tocdepth}{3}}

\section{Additional Implementation Details}
\label{sec:supp_detail}

In this section, we provide additional implementation details for the baseline models and the No-thinking methods discussed in the main paper.

\subsection{Implementation Details on Baseline}
\textbf{Seg-Zero Re-implementation.} 
The evaluation of Seg-Zero~\cite{liu2025seg}'s 7B model  is conducted using the official model checkpoint available in the public repository.
Due to the unavailability of the 3B checkpoint, we re-train the model using the official codebase under the same experimental settings to ensure fair comparison.
The prompts used for both models strictly adhere to the official implementation, as shown below. 
For consistency, the same prompt format is also adopted in our proposed \model.

\begin{tcolorbox}[title=Original Prompt from Seg-Zero,colback=blue!10!white,colframe=violet!60!black,fonttitle=\bfseries]
Please find ``\textcolor{blue}{\{Question\}}'' with bbox and points. 

Compare the differences between objects and find the most closely matched one. 

Output the \textit{thinking process} in \texttt{<think> </think>} and \textit{final answer} in \texttt{<answer> </answer>} tags. 

Output the \textit{one bbox} and \textit{points} of two largest inscribed circles inside the interested object in JSON format. 

\textit{i.e.}, \texttt{<think>} \textit{thinking process here} \texttt{</think>} 

\texttt{<answer>}{{``Bbox'': $[10,100,200,210]$, ``Point $1$'': $[30,110]$, ``Point $2$'': $[35,180]$}}\texttt{</answer>}
\end{tcolorbox}

\textbf{Seg-Zero with Prompt for Short-Thinking.}
For the prompt-based baseline, we utilize the official model checkpoint and explicitly \textit{incorporate a token budget constraint into the prompt}, setting the upper limit to $64$ tokens during inference. 
The specific prompt used is provided below:

\begin{tcolorbox}[title=Short-thinking Prompt,colback=blue!10!white,colframe=violet!60!black,fonttitle=\bfseries]
Please find ``\textcolor{blue}{\{Question\}}'' with bbox and points. 

Compare the difference between objects and find the most closely matched one. 

Think step-by-step and explain your reasoning process in less than $64$ tokens. 

Output the \textit{reasoning} in  \texttt{<think> </think>} and the \textit{final answer} in \texttt{<answer> </answer>} tags.

Output the \textit{one bbox} and \textit{points} of two largest inscribed circles inside the interested object in JSON format. 

\textit{i.e.}, \texttt{<think>} \textit{short reasoning here} \texttt{</think>} 

\texttt{<answer>}{{``Bbox'': $[10,100,200,210]$, ``Point $1$'': $[30,110]$, ``Point $2$'': $[35,180]$}}\texttt{</answer>}
\end{tcolorbox}

\textbf{Adapt L1 for Reasoning Segmentation.}
Since L1~\cite{aggarwal2025l1} is designed to control the reasoning length of large language models (LLMs)~\cite{guo2025deepseekr1,yang2024qwenllm}, its original implementation requires explicitly specifying the desired token count in the prompt. 
Besides, it is initially trained on the fixed-length DeepScaleR dataset~\cite{luo2025deepscaler} using L1-Exact, followed by continued fine-tuning with L1-Max. 
To distinguish L1 from prompt-based baseline and ensure a fair comparison with Seg-Zero without relying on \textit{additional reasoning data}, we adopt the same prompt format used in Seg-Zero for re-implementing L1 in the reasoning segmentation task. 
For both variants, we independently integrate the corresponding length control functions into the reward formulation. Specifically, L1-Exact uses a $64$-token upper limit, while L1-Max allows up to $128$ tokens, enabling comparable final reasoning lengths.

\subsection{Implementation Details on No-thinking Mode}
For the implementation of No-thinking-Prompt, we directly use the official model checkpoint from Seg-Zero and apply a prompt that explicitly instructs the model not to produce any reasoning steps as below, thereby enforcing a ``no-thinking'' behavior. 
For No-thinking-RL, we follow the CLS-RL~\cite{li2025cls} framework by adopting a similar prompt format and removing the reasoning-format reward term from the GRPO reward function. 
We then re-train Seg-Zero using this modified reward formulation to obtain the final No-thinking-RL results.

\begin{tcolorbox}[title=No-thinking Prompt,colback=blue!10!white,colframe=violet!60!black,fonttitle=\bfseries]
Please find ``\textcolor{blue}{\{Question\}}'' with bbox and points.

Compare the differences between objects and find the most closely matched one. 

Output the final answer in the \texttt{<answer> </answer>} tag only. 

The answer should include \textit{one bbox} and the \textit{points} of the two largest inscribed circles inside the interested object in JSON format, \textit{i.e.},

 \texttt{<answer>}{{``Bbox'': $[10,100,200,210]$, ``Point $1$'': $[30,110]$, ``Point $2$'': $[35,180]$}}\texttt{</answer>}
\end{tcolorbox}
\section{The \dataset Dataset}
\label{sec:supp_dataset}

In this section, we present the dataset construction details, including the scoring prompts and representative examples from \dataset.

\subsection{Prompts for Scoring and Statistics}
\noindent \textbf{Difficulty Scoring.} 
For both the RefCOCOg~\cite{yu2016refcoco} training set and the validation/test splits constructed from ReasonSeg~\cite{lai2024lisa} to form \dataset, we need to assign a difficulty score for each sample. 
To achieve this, we generate \textit{visual descriptions} based on mask properties (\textit{e.g.}, size and position) and \textit{textual descriptions} derived from the referring expressions (\textit{e.g.}, expression length and the number of spatial terms). 
These descriptions are incorporated into the prompt.
We then instruct Qwen2.5-VL-72B~\cite{bai2025qwen25} to rate each sample from three perspectives: (1) \textit{Scene Complexity}, (2) \textit{Segmentation Challenge}, and (3) \textit{Linguistic Ambiguity}. 
The final difficulty score is computed as the average of these three ratings. The prompt used for this scoring process is provided below.

\begin{tcolorbox}[title=Difficulty Scoring Prompt,colback=blue!10!white,colframe=violet!60!black,fonttitle=\bfseries]
You are an expert in reasoning segmentation evaluation.
 
Given the image and the referring expression: ``\textcolor{blue}{\{Question\}}'', please assess the task difficulty based on the following three aspects:

1. \textit{Scene Complexity}:- How many objects are visible in the scene?- How many of them are potentially related or visually similar to the target?

2. \textit{Segmentation Challenge}:

- What is the size and position of the target object?

- Are there occlusions, overlaps, or visually similar objects nearby?

- Is the mask describing the whole object or just a part?
``\textcolor{blue}{\{Visual Description\}}''

3. \textit{Language Complexity}:

- Does the referring expression explicitly point to the target object?

- Or does it require additional reasoning to infer which object is referred to?
``\textcolor{blue}{\{Textual Description\}}''

For each aspect, please provide a difficulty rating from $1$ (very easy) to $10$ (very hard), and summarize in the following Python dictionary format.

\textit{i.e.}, \{``scene'': $4$, ``segmentation'': $6$, ``language'': $3$\}
\end{tcolorbox}

\begin{wraptable}{r}{0.48\linewidth}
    \centering
    \captionsetup{width=.48\textwidth}
    \caption{Label statistics on difficulty distribution in training, validation and test set.}
    \vspace{-2mm}
    \label{tab:supp_stat}
    \resizebox{0.48\textwidth}{!}{
    \begin{tabular}{c|ccc}
    \toprule[1.25pt]
    \rowcolor{gray!11} Dataset & Easy  & Medium & Hard 
    \\\midrule\midrule
    Training Set (RefCoCOg)  &  $3220$  & $4810$ & $970$  
    \\
    Validation Set (ReasonSeg)   &  $51$   & $102$ & $45$ 
    \\
    Test Set (ReasonSeg)   &  $171$   & $411$ & $187$ 
    \\

    \bottomrule
    \end{tabular}}
    \vspace{-2mm}
\end{wraptable}
\noindent \textbf{Lable Statistics.} 
We further analyze the distribution of difficulty levels, namely \textit{easy}, \textit{medium}, and \textit{hard}, as determined by our scoring framework for both RefCOCOg~\cite{yu2016refcoco} and ReasonSeg~\cite{lai2024lisa}. 
As summarized in Table~\ref{tab:supp_stat}, RefCOCOg contains a significantly higher proportion of easy samples compared to ReasonSeg, which is consistent with commonly held expectations regarding the relative complexity of the two datasets. 
Notably, despite differences in data distribution between RefCOCOg and ReasonSeg, our method achieves superior \textit{zero-shot} performance on ReasonSeg when trained solely on RefCOCOg.

\noindent \textbf{Reasoning Process Scoring.}
For reasoning process evaluation, we adopt the following prompt and use reasoning chains from \dataset  as reference annotations. 
The model-generated reasoning is evaluated by Qwen2.5-72B~\cite{yang2024qwenllm} across three dimensions: \textit{Completeness}, \textit{Object grounding}, and \textit{Fluency \& Clarity}, providing a comprehensive assessment of reasoning quality.

\begin{tcolorbox}[title=Reasoning Scoring Prompt,colback=blue!10!white,colframe=violet!60!black,fonttitle=\bfseries]
You are an expert in evaluating reasoning quality for reasoning segmentation tasks. 

Given the following predicted reasoning and reference reasoning, please score the prediction in three aspects from 1 to 10:

1. \textit{Completeness}: Does it include all necessary steps and important information?

2. \textit{Object Grounding}: Is it aligned with the referred object in the question?

3. \textit{Fluency \& Clarity}: Is the reasoning coherent, fluent, and grammatically correct?

The question is: ``\textcolor{blue}{\{Question\}}''

Reference Reasoning: ``\textcolor{blue}{\{Reference Text\}}''

Predicted Reasoning: ``\textcolor{blue}{\{Thinking Text\}}''

Return a Python dictionary with keys ``completeness'', ``grounding'', and ``fluency''. 

\textit{i.e.}, \{``completeness'': $8$, ``grounding'': $7$, ``fluency'': $9$\}
\end{tcolorbox}

\subsection{Examples from \dataset}

\begin{figure}[ht]
    \centering
    \includegraphics[width=\linewidth]{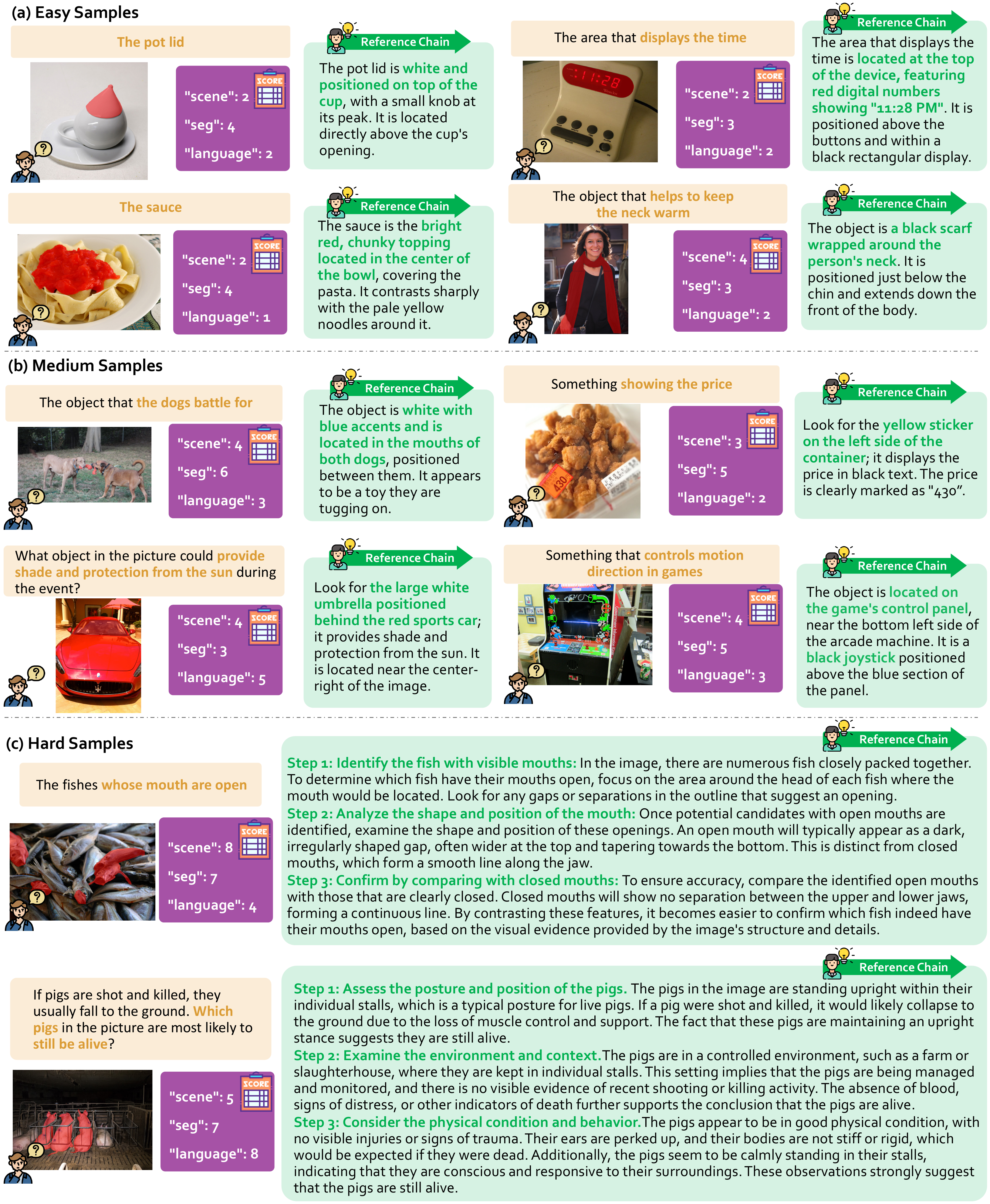}
    \caption{\textbf{Qualitative examples from \dataset.} Representative samples are shown with annotated difficulty scores and corresponding reference reasoning chains. For easy and medium cases, short reasoning chains are preferred to capture concise visual cues, whereas hard cases benefit from longer chains that reflect more elaborate reasoning over ambiguous or complex scenes. These examples illustrate the diverse reasoning requirements across difficulty levels and support more fine-grained evaluation.}
    \label{fig:supp_examples}
\end{figure}

We present several representative examples from the constructed \dataset in Figure~\ref{fig:supp_examples}, covering samples categorized as \textit{easy}, \textit{medium}, and \textit{hard}. Each example includes its assigned difficulty scores along with the corresponding reference reasoning chains. For \textit{easy} and \textit{medium} cases, we recommend the use of \textit{short} reasoning chains, whereas \textit{longer} chains are preferable for \textit{hard} samples. The annotation files are available in the \dataset directory  as a \textit{supplementary attachment} for further reference.

\subsection{License}
The \dataset dataset is released under the Attribution-ShareAlike 4.0 International (CC BY-SA 4.0)\footnote{\url{https://creativecommons.org/licenses/by-sa/4.0/legalcode}.} license.

\section{Additional Experimental Results}
\label{sec:supp_exp}

In this section, we conduct more ablation experiments and provide additional qualitative results including failure cases in reasoning segmentation.

\subsection{Additional Ablation Results}
\label{sec:more_abl}
All experiments in this section follow the main ablation setting, using the 7B model and evaluating on the validation split.

\begin{table}[t]
    \centering
    \begin{minipage}[t]{0.45\textwidth}
    \centering
    \caption{Ablation on the uncertainty weight.}
    \label{tab:abl_unc}
    \resizebox{\textwidth}{!}{
    \begin{tabular}{c|ccc}
    \toprule[1.25pt]
    \rowcolor{gray!11}Weight ($\alpha$)  & \#Token $\downarrow$  & \texttt{gIoU}(\%)$\uparrow$ & \texttt{cIoU}(\%)$\uparrow$ 
    \\\midrule\midrule
    $0$  &  $\mathbf{39.95}$  & $62.07$ & $58.35$  
    \\
    $25$   &  $46.98$   & $\mathbf{63.81}$ & ${62.69}$ 
    \\
    $35$  &  $71.10$    & $62.66$ & $\mathbf{63.13}$ 
    \\
    \bottomrule
    \end{tabular}
    }
    \end{minipage}
    \hfill
    \begin{minipage}[t]{0.45\textwidth}
    \centering
    \caption{Ablation on the length constrain for \textit{medium} samples during training.}
    \label{tab:abl_middle}
    \resizebox{\textwidth}{!}{
    \begin{tabular}{c|ccc}
    \toprule[1.25pt]
    \rowcolor{gray!11} \textit{with} constrain & \#Token $\downarrow$  & \texttt{gIoU}(\%)$\uparrow$ & \texttt{cIoU}(\%)$\uparrow$ 
    \\\midrule\midrule
    \cmark  &  $\mathbf{42.10}$  & $60.92$ & $53.84$  
    \\
    \xmark   &  $46.98$   & $\mathbf{63.81}$ & $\mathbf{62.69}$ 
    \\

    \bottomrule
    \end{tabular}
    }
    \end{minipage}
    \vspace{-3mm}
\end{table}

\noindent \textbf{Ablation on Uncertainty Weight.}
We conduct an ablation study on the uncertainty ($\mathcal{U}$) weighting factor $\alpha$ for hard samples to evaluate its effect on reasoning length control and segmentation performance. 
As shown in Table~\ref{tab:abl_unc}, selecting an appropriate uncertainty weight is crucial for achieving optimal performance in both reasoning efficiency and segmentation accuracy.

\noindent \textbf{Ablation on Length Constrain for Medium Samples.}
We also investigate the impact of applying length constrain to medium-difficulty samples. 
In the controlled variant, the reasoning length is limited to a maximum of $176$ tokens.
As shown in Table~\ref{tab:abl_middle}, allowing medium cases to remain unconstrained provides greater flexibility in reasoning length, resulting in improved overall performance.

\subsection{Additional Qualitative Results}

\begin{figure}[t]
    \centering
    \includegraphics[width=\linewidth]{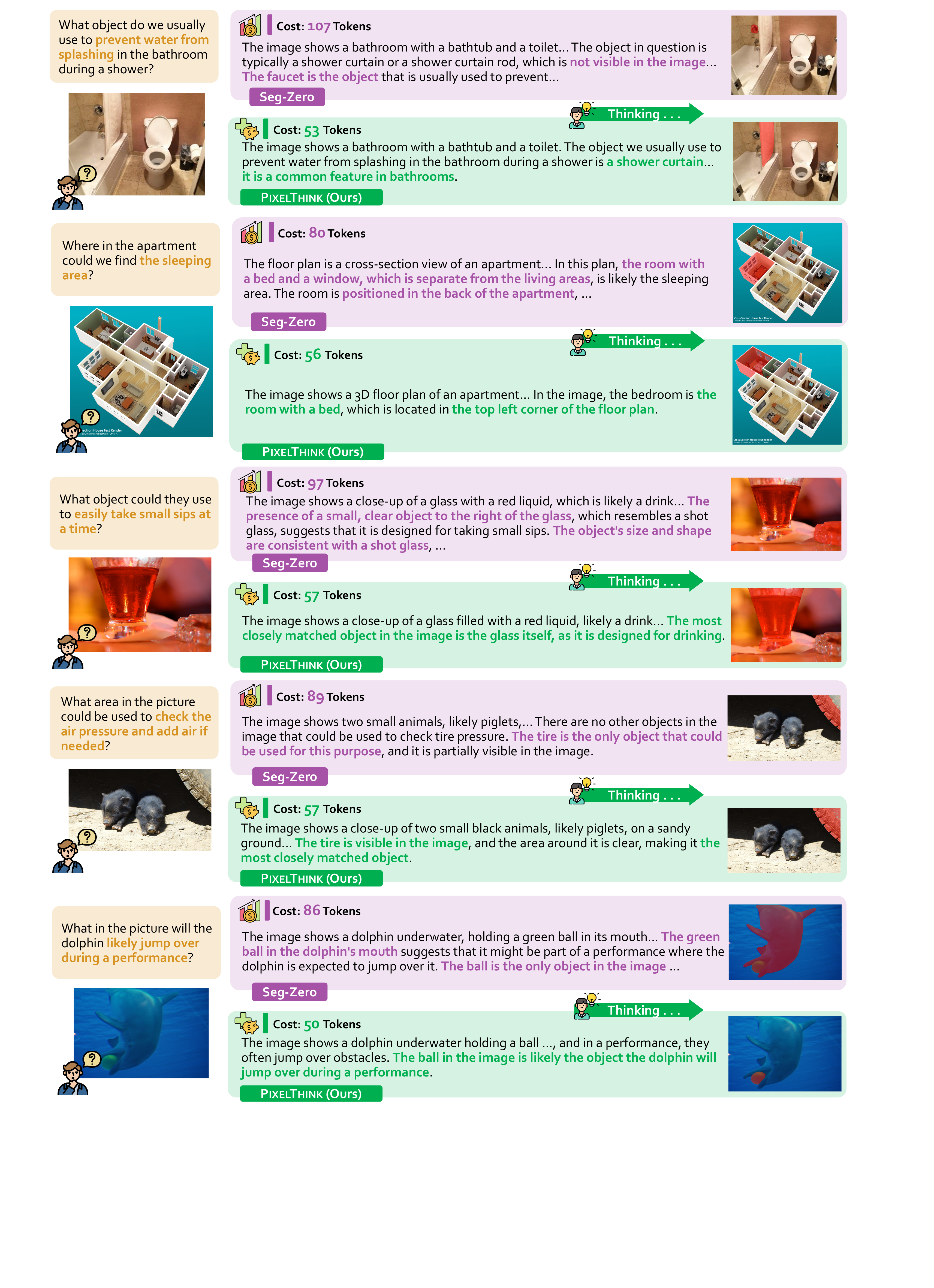}
    \caption{Additional \textbf{qualitative comparisons} between Seg-Zero~\cite{liu2025seg} and the proposed \model. Our method demonstrates consistent improvements in segmentation accuracy across diverse scenarios, accompanied by substantially shorter reasoning chains. These examples highlight the effectiveness of our efficient reasoning framework in mitigating \textit{overthinking} while maintaining or improving segmentation quality.}
    \label{fig:supp_vis}
\end{figure}

In Figure~\ref{fig:supp_vis}, we provide additional qualitative comparisons between our \model and Seg-Zero, including both segmentation masks and reasoning chains. Across a range of scenarios, \model consistently yields more accurate segmentation results while generating significantly shorter reasoning chains, highlighting its superior \textit{efficiency} and \textit{effectiveness}.

\subsection{Failure Cases and Analyses}

\begin{figure}[t]
    \centering
    \includegraphics[width=\linewidth]{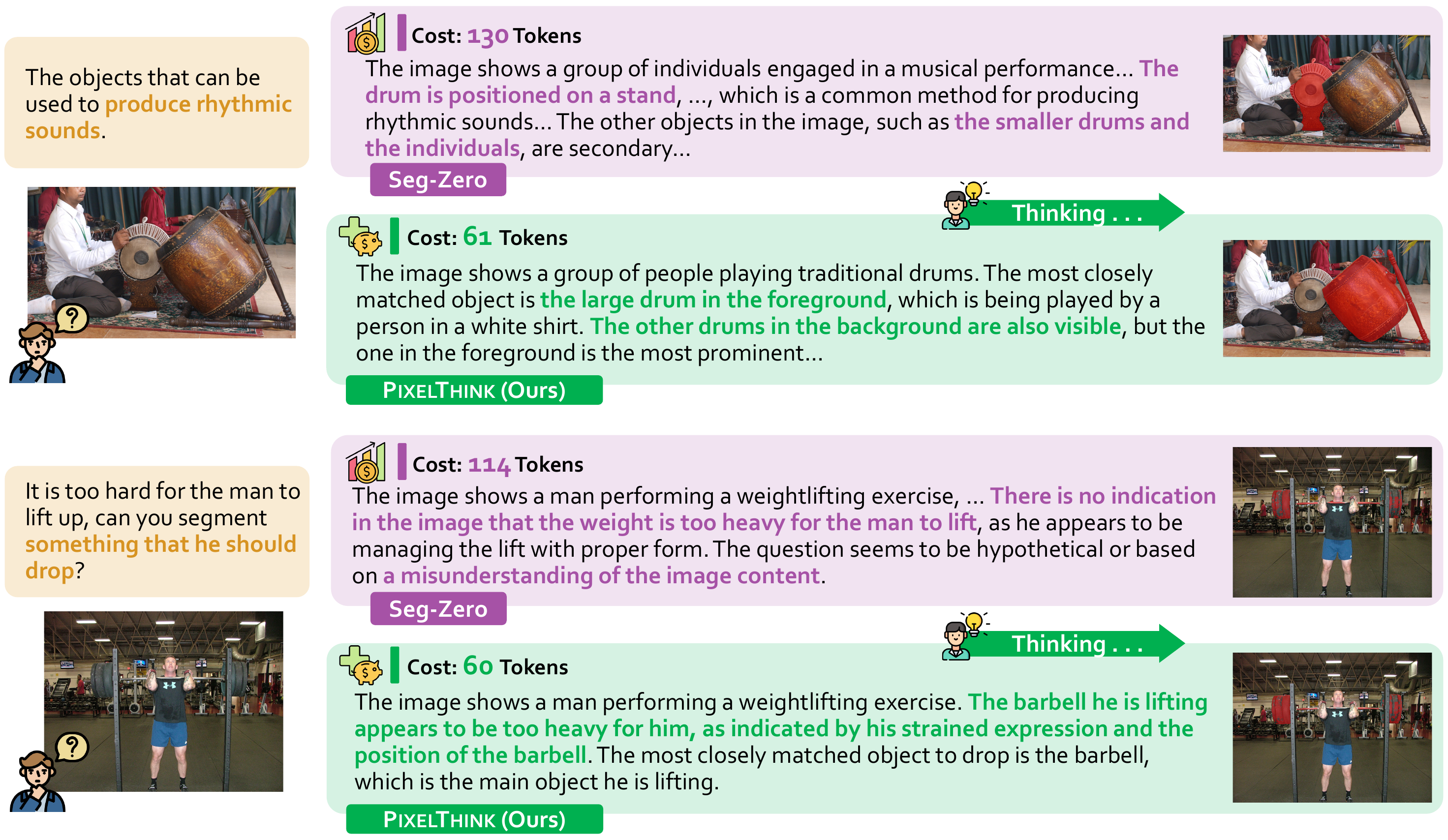}
    \caption{\textbf{Failure cases} of Seg-Zero~\cite{liu2025seg} and the proposed \model. The first example illustrates ambiguity in the referring expression, resulting in incomplete segmentation masks from both methods. The second example reveals a mismatch between reasoning and segmentation: Seg-Zero fails to identify the correct object in its reasoning, while \model correctly interprets the instruction but produces a partially inaccurate mask.}
    \label{fig:failure_case}
\end{figure}

In Figure~\ref{fig:failure_case}, we present several failure cases to illustrate limitations of the current approach. In the first example, both Seg-Zero and \model produce incomplete segmentation results, as multiple objects in the scene satisfy the referring conditions. In the second example, Seg-Zero fails during the reasoning process by misidentifying the target object, yet still generates a partially correct mask. 
In contrast, our method correctly identifies the target but yields an imperfect segmentation mask. These cases reveal a key limitation of the current decoupled architecture, where the reasoning outputs and final masks are not always well aligned. 
In the future, we will explore tighter integration and joint optimization to enhance consistency and overall performance in reasoning segmentation.

\section{Additional Observations and Analyses}
\label{sec:supp_further}

In this section, we provide complementary observations to further interpret the experimental results presented in the main paper. Specifically, we examine several empirical phenomena observed during training and evaluation: 
(1) the discrepancy between the token budget and the actual reasoning length, 
(2) the convergence of reasoning lengths across varying difficulty levels, 
and (3) the trade-off between concise reasoning and completeness in comparison to Seg-Zero.

\subsection{Discrepancy between Token Budget and Reasoning Length}
In our experiments, we observe that the model trained with the proposed reward framework frequently generates reasoning chains that are significantly shorter than the predefined token budget. This behavior can be attributed to several factors:

\noindent \textbf{Soft Length Penalty Encourages Conservative Generation.}
Our reward function incorporates a \textit{soft length penalty} that linearly penalizes the use of tokens exceeding the expected budget. 
Unlike hard truncation, this approach allows for flexibility while implicitly encouraging the model to stay within budget. 
Therefore, the model learns to \textit{avoid unnecessary token usage} unless it contributes to improved task performance.

\noindent \textbf{Accuracy-dominant Reward Prevents Token Inflation.}
The final reward integrates segmentation accuracy with alignment to the expected reasoning length. 
Since $\mathcal{R}_{\mathrm{original}}$ primarily governs the reward dynamics, longer reasoning chains that do not lead to performance gains are \textit{implicitly penalized}. 
This design encourages the generation of concise yet informative reasoning, where token usage is closely aligned with task utility.

\subsection{Convergence of Reasoning Length across Difficulty Levels}
Although our training framework assigns distinct token budgets according to difficulty levels, we notice that the final reasoning lengths across all categories tend to converge within a relatively narrow range. 
Several determinants account for this phenomenon:

\noindent \textbf{Shared Decoder and Autoregressive Generation Bias.}
The reasoning model employs a \textit{unified decoder} with \textit{same prompt} to generate reasoning chains for all samples. 
Since this decoder is optimized across tasks with varying levels of difficulty, it learns an averaged generation pattern and tends to favor a stable reasoning length distribution. 
This behavior is further reinforced by the autoregressive nature of decoding, through which the model implicitly learns a preferred stopping condition based on distributional patterns observed during training.

\noindent \textbf{Conservative Token Budget Design.}
The token budget upper bounds for each difficulty group are set conservatively high to avoid premature truncation. The soft penalty is applied only when the reasoning length exceeds the budget and does not actively encourage the model to approach the upper limit. 
This design allows the model to naturally converge to a reasoning length below the threshold.

\subsection{\texttt{RScore} Performance Compared to Seg-Zero}
While our method surpasses Seg-Zero in both segmentation accuracy and inference efficiency, we observe slightly lower values in the \texttt{RScore}, which evaluates the quality of the generated reasoning chains. 
This can be traced to limitations in the design of the \texttt{RScore} metric:

\noindent \textbf{\texttt{RScore} Emphasizes Completeness without Length Awareness.}
\texttt{RScore} is computed based on three criteria: \textit{completeness}, \textit{grounding}, and \textit{fluency}. Notably, the metric does not consider the brevity or efficiency of the generated reasoning. 
Thus, longer reasoning chains often receive higher completeness scores, even when parts of the explanation may be redundant.

\noindent \textbf{\model Prioritizes Efficiency and Accuracy.}
\model is designed to generate concise yet informative reasoning under length-aware constraints. While our reasoning is more efficient, it may omit minor details which can lead to slightly lower completeness scores. However, these omissions \textit{do not necessarily affect segmentation accuracy}, which remains higher in our approach.

\noindent \textbf{\texttt{RST} Facilitates a More Equitable Evaluation.}
To address this limitation, we propose \textbf{Reasoning Score per Token} (\texttt{RST}), which normalizes \texttt{RScore} by both model size and the number of generated tokens. 
This metric offers a more holistic assessment of reasoning quality relative to computational cost, enabling fairer comparisons between models with varying reasoning lengths.

\section{Further Discussions}
\label{sec:supp_dis}
In this section, we further discuss the limitations of our work, highlight directions for future research and consider the potential societal impact.

\subsection{Limitation and Future Work}
\label{sec:supp_dis_limit}
As the first attempt to enable efficient reasoning in reasoning segmentation, our method emphasizes simplicity and practicality, focusing on token-level control guided by task difficulty and uncertainty. However, our design still relies on coarse-grained difficulty scores and manually defined budget rules, which may limit adaptiveness in more complex scenarios. Additionally, the reasoning and segmentation stages are loosely coupled, and the applicability of our framework to broader multimodal tasks remains to be fully validated.

In the future, we will explore more precise and robust difficulty estimation by leveraging self-supervised signals in combination with human feedback, as well as developing finer-grained, learnable token allocation strategies that adapt to task-specific demands. 
Furthermore, integrating reasoning and segmentation into a joint optimization framework improves consistency and overall performance. 
Extending the proposed paradigm to other vision-language reasoning tasks such as visual question answering (VQA), visual commonsense reasoning, and video understanding further demonstrates its generalizability and practical value.

\subsection{Potential Societal Impact}
\label{sec:supp_dis_social}
In this work, a reinforcement learning-based fine-tuning scheme is proposed for efficient reasoning in segmentation tasks, with potential applications in domains such as autonomous driving, robotics, and medical imaging. 
By enabling more efficient and interpretable visual reasoning, our method supports safer and more transparent decision-making in high-stakes scenarios. However, as with many vision-language models, our approach depends on large-scale pretrained models, which may carry biases from their training data. Moreover, automated reasoning systems could be misapplied in contexts such as surveillance or critical decision-making without adequate human oversight. We advocate for responsible deployment and encourage further research on fairness, robustness, and transparency to ensure beneficial societal impact.

\clearpage\clearpage
\section{License and Consent with Public Resources}
\label{sec:supp_license}
In this section, we outline the details of responsible release and acknowledge the use of public resources that supported this work.

\subsection{Responsible Release}
Our work focuses on enhancing efficiency and controllability in reasoning segmentation using publicly available models and datasets. The proposed benchmark, \dataset, is constructed based on ReasonSeg~\cite{lai2024lisa}, which includes human-annotated referring expressions and segmentation masks from natural scenes. We ensure that no private, sensitive, or copyrighted content is introduced during data processing. All models used are open-sourced under appropriate licenses, and we do not release any newly trained large-scale models that could pose potential misuse risks. Upon releasing our benchmark and codebase, we will include a clear license, data usage policy, and guidelines to discourage applications involving sensitive attributes, surveillance, or unauthorized identification.

\subsection{Public Datasets}
All experiments and the construction of our benchmark are conducted using the following publicly available datasets:
\begin{itemize}
    \item ReasonSeg~\cite{lai2024lisa}\footnote{\url{https://github.com/dvlab-research/LISA}.} \dotfill Apache License 2.0
    \item RefCOCO (+/g)~\cite{yu2016refcoco}\footnote{\url{https://github.com/lichengunc/refer}.} \dotfill Apache License 2.0
    \item MS COCO~\cite{lin2014microsoft}\footnote{\url{https://cocodataset.org}.} \dotfill Other (specified in description)
\end{itemize}

\subsection{Public Models and Implementation}
We compare and validate the effectiveness of the proposed method using the following publicly available models and source codes:
\begin{itemize}
     \item Qwen2.5-VL~\cite{bai2025qwen25}\footnote{\url{https://github.com/QwenLM/Qwen2.5-VL}.} \dotfill Apache License 2.0
     \item SAM2~\cite{ravi2024sam2}\footnote{\url{https://github.com/facebookresearch/sam2}.} \dotfill Apache License 2.0
     \item Seg-Zero~\cite{liu2025seg}\footnote{\url{https://github.com/dvlab-research/Seg-Zero}.} \dotfill Apache License 2.0
     \item verl~\cite{sheng2024hybridflow}\footnote{\url{https://github.com/volcengine/verl}.} \dotfill Apache License 2.0
     \item L1~\cite{bai2025qwen25}\footnote{\url{https://github.com/cmu-l3/l1}.} \dotfill MIT License
\end{itemize}

\clearpage\clearpage
\bibliographystyle{unsrt}
{
\small
\bibliography{main.bib}
}

\end{document}